\documentclass[letterpaper, 10 pt, conference]{ieeeconf}  

\IEEEoverridecommandlockouts                              
\usepackage{amsmath,amssymb,bm}
\usepackage{array}
\usepackage{graphicx}
\usepackage{color}
\usepackage[table]{xcolor}
\usepackage{lscape}
\usepackage{multirow}
\usepackage{cite}
\usepackage{float}
\usepackage{url}
\usepackage{soul}
\usepackage{longtable}
\usepackage{booktabs}
\usepackage{tikz}
\usetikzlibrary{shapes.geometric}  
\usepackage{balance}

\title{\LARGE \bf
Koopman Operator Identification of Model Parameter Trajectories for Temporal Domain Generalization (KOMET)
}

\author{Randy C. Hoover, Jacob James, Paul May, and Kyle Caudle
\thanks{R. C. Hoover and P. May are with the Department of Electrical Engineering and Computer Science (EECS), South Dakota Mines, Rapid City, SD 57702, USA
        {\tt\small \{randy.hoover, paul.may\}@sdsmt.edu}}%
\thanks{J. James is a Ph.D. student in the Data Science and Engineering Program within EECS, South Dakota Mines, Rapid City, SD 57702, USA
        {\tt\small jacob.james@mines.sdsmt.edu}}%
\thanks{K. Caudle is with the Department of Mathematics, South Dakota Mines, Rapid City, SD 57702, USA
        {\tt\small kyle.caudle@sdsmt.edu}}%
}

\begin{document}

\maketitle
\thispagestyle{empty}
\pagestyle{empty}

\begin{abstract}

Parametric models deployed in non-stationary environments degrade as the
underlying data distribution evolves over time (a phenomenon known as
temporal domain drift).  In the current work, we present \textbf{KOMET} (Koopman Operator
identification of Model parameter Evolution under Temporal drift), a
model-agnostic, data-driven framework that treats the sequence of trained
parameter vectors as the trajectory of a nonlinear dynamical system and
identifies its governing linear operator via Extended Dynamic Mode
Decomposition (EDMD).  A warm-start sequential training protocol enforces
parameter-trajectory smoothness, and a Fourier-augmented observable
dictionary exploits the periodic structure inherent in many real-world
distribution drifts.  Once identified, KOMET's Koopman operator predicts
future parameter trajectories \emph{autonomously}, without access to
future labeled data, enabling zero-retraining adaptation at deployment.
Evaluated on six datasets spanning rotating, oscillating, and expanding
distribution geometries, KOMET achieves mean autonomous-rollout
accuracies between 0.981 and 1.000 over 100 held-out time steps.  Spectral and coupling analyses further reveal
interpretable dynamical structure consistent with the geometry of the
drifting decision boundary.
\end{abstract}

\section{Introduction}
\label{sec:introduction}

Parametric models deployed in real-world environments routinely encounter data
whose statistical properties drift over time, a phenomena collectively described
as \emph{distribution drift} or \emph{concept drift}~\cite{lu2018learning,gama2014survey}.
The machine-learning community has addressed this primarily through
\emph{Domain Generalization} (DG), where they train on source domains and generalize to
an unseen targets~\cite{wang2022generalizing,gulrajani2021search}, using
techniques such as domain-adversarial learning~\cite{ganin2016domain},
invariant risk minimization~\cite{arjovsky2019invariant}, transfer learning~\cite{9134370}, and distributionally
robust optimization~\cite{sagawa2020distributionally}.  These methods share a
critical limitation because domains are treated as unordered, so temporal structure
carries no predictive information about the future.

\emph{Temporal Domain Generalization} (TDG) removes this restriction by leveraging dynamical systems theory to 
model the ordered domain sequence as a time series~\cite{yao2022wildtime}.
Within TDG, the central question is how to represent and exploit the dynamics
of change.  DRAIN~\cite{bai2023drain} represents each model snapshot as a
dynamic graph and trains a Bayesian recurrent generator to extrapolate full
weight sets to the next unseen domain.  Koodos~\cite{cai2024koodos} extends
this to continuous time, constructing a neural-ODE Koopman embedding over the
joint space of distributions and parameters.  Complementary approaches
track evolving normalization statistics~\cite{xie2024evolving}, enforce
continuous invariance~\cite{lin2024continuous}, or apply incremental
online-learning strategies~\cite{losing2018incremental,zenke2017continual}.  A concurrent complementary approach, SLATE~\cite{james2026slate}, addresses the same TDG setting via a low-rank batch-smoothing factorization of the full parameter trajectory, using cubic B-spline temporal bases and an AR(1) Gauss-Markov prior to enforce smoothness without recurrence or sequential rollout. The current work and SLATE thus target the same problem from orthogonal angles, inverse dynamical-systems identification versus structured trajectory smoothing.

A parallel thread has matured in dynamical systems via
\emph{Koopman operator theory}~\cite{koopman1931hamiltonian,mezic2005spectral}.
The Koopman operator is an infinite-dimensional linear operator governing the
evolution of observable functions of a nonlinear system, enabling exact global
linearization of complex dynamics.  Practical finite-dimensional approximations
are computed by \emph{Extended Dynamic Mode Decomposition}
(EDMD)~\cite{williams2015data}, which fits a linear operator in a
user-specified observable dictionary where its eigenvalues directly encode dominant
frequencies and decay rates~\cite{brunton2022modern}.  Deep learning has been
combined with EDMD to learn dictionaries
automatically~\cite{lusch2018deep,morton2018deep}, and the connection to
neural ODEs has been clarified~\cite{otto2023koopman}.

Despite rapid advances in both TDG and Koopman methods, a fundamental
perspective is absent: treating the sequence of \emph{trained parameter
vectors themselves} as a nonlinear dynamical system and performing rigorous
system identification on observed trajectories.  DRAIN and Koodos answer a
forward prediction question which is, given a domain history, predict weights for the
next domain, however, neither recovers the mechanistic law governing how parameters
evolve, nor exposes the spectral modes and coupling structure that enable
analytical characterization of the drift.

\subsection{Contributions of the Current Work}

We propose \text{KOMET} (\text{Ko}opman identification of \text{M}odel
parameter \text{E}volution under \text{T}emporal drift), a model-agnostic
pipeline for post-hoc dynamical-systems identification of parametric model
parameter trajectories under periodic distribution drift.  KOMET treats the model itself as an underlying dynamical system and is
\emph{complementary} to DRAIN and Koodos in that, they solve a forward
prediction problem inside a joint training loop, KOMET solves an inverse
identification problem on observed model parameter trajectories, requiring no architectural
modification and no distributional model of the drift.  Specifically, the current paper provides the following contributions:

\subsubsection{EDMD Koopman identification and autonomous rollout}
A physics-informed dictionary (Fourier harmonics at the known drift period plus
polynomial terms in the PCA-whitened state) is used to fit a low-dimensional
Koopman operator via EDMD, then rolled out autonomously over the held-out horizon
without access to future labeled training data.

\subsubsection{Exact forward-prediction performance}
On all datasets satisfying the periodicity assumption, autonomous prediction
matches the fully-retrained upper bound (mean accuracy $0.981$--$1.000$, zero
timesteps below $90\%$ over 100 held-out steps) across binary and three-class
tasks spanning diverse drift geometries.

\subsubsection{Interpretable spectral and coupling analysis}
The Koopman eigenspectrum encodes drift periodicity and attractor damping.
Distance correlation and transfer entropy analyses expose layer-wise coupling demonstrating that first-layer weights dominate under simple drift (TE ratio $1.95\times$
layer L1$\to$ layer L2) and become bidirectional as classification difficulty increases,
consistent with harder representational demands at Layer~1.

\subsubsection{Warm-start training with Adam moment continuity}
Parameters are trained sequentially across $T$ timesteps, warm-starting both
weights and the full Adam state across transitions.  A temporal smoothness
penalty $\lambda_s\|\bm\theta_t - \hat{\bm\theta}_{t-1}\|^2$, where $\hat{\bm\theta}$ is a fixed estimate of the the parameter ${\bm\theta}$, is added to the loss to prevent trajectory
discontinuities. For softmax networks, $\ell_2$ weight decay counteracts
scale-invariance that would otherwise drive unbounded weight growth and violate
the Koopman periodicity assumption and stability enforcement through spectral reprojection.  Together these reduce mean
epochs-to-convergence by $9$-fold and produce smooth, Koopman-amenable weight trajectories.

The remainder of this paper is organized as follows: Section~\ref{sec:background} provides some background material and mathematical preliminaries required to understand the KOMET pipeline.  Section~\ref{sec:method} outlines the entire KOMET approach and the two phase pipeline.  Section~\ref{sec:experiments} outlines the datasets used for experimental analysis within the KOMET framework and presents experimental results compared to baseline retraining or frozen-start methods.  Finally, Section~\ref{sec:conclusion} outlines some conclusions and provides some promising future directions to explore.


%
\section{Background \& Mathematical Preliminaries}
\label{sec:background}

\subsection{Koopman Operator Theory}
\label{ssec:koopman}

Consider a discrete-time autonomous dynamical system
\begin{equation}
  \bm\theta_{t+1} = F(\bm\theta_t), \quad \bm\theta_t \in \mathcal{M} \subseteq \mathbb{R}^{n_\theta},
  \label{eq:dynamics}
\end{equation}
where $F:\mathcal{M}\to\mathcal{M}$ is in general nonlinear.  The
\emph{Koopman operator} $\mathcal{K}$~\cite{koopman1931hamiltonian} is an infinite-dimensional linear
operator that acts on scalar \emph{observable} functions $g:\mathcal{M}\to\mathbb{R}$
according to
\begin{equation}
  \mathcal{K}\,g = g \circ F.
  \label{eq:koopman}
\end{equation}
Linearity is the key property because even when $F$ is nonlinear, \emph{every}
observable evolves linearly under $\mathcal{K}$.  The tradeoff is
dimensionality, i.e., the operator acts on a function space rather than on
$\mathcal{M}$ directly.  If a finite-dimensional \emph{Koopman-invariant
subspace} exists, spanned by observables $\{\psi_1,\ldots,\psi_N\}$, then the
dynamics of the lifted state vector
$\bm\psi(\bm\theta) = [\psi_1(\bm\theta),\ldots,\psi_N(\bm\theta)]^\top$
are governed exactly by a finite matrix $\mathbf{A}\in\mathbb{R}^{N\times N}$ such that
\begin{equation}
  \bm\psi(\bm\theta_{t+1}) = \mathbf{A}\,\bm\psi(\bm\theta_t).
  \label{eq:koopman_finite}
\end{equation}
The eigenvalues $\{\lambda_i\}$ of $\mathbf{A}$, the \emph{Koopman eigenvalues},
encode temporal frequencies and growth rates of the underlying dynamics.
For a stable autonomous system the spectral radius $\rho(\mathbf{A})\leq 1$ ensures
that autonomous rollouts (i.e., forward propagation) of~\eqref{eq:koopman_finite} are forward invariant.  Mezić~\cite{mezic2005spectral} established the spectral foundations,
Brunton et al.~\cite{brunton2022modern} provide a comprehensive modern review, and
the Koopman spectral approach to control is present in~\cite{goswami2017bilinearization}.

\subsection{Extended Dynamic Mode Decomposition}
\label{ssec:edmd}

Extended Dynamic Mode Decomposition (EDMD)~\cite{williams2015data} fits a
finite-dimensional approximation of $\mathbf{A}$ from data.  Given a
trajectory $\{\bm\theta_t\}_{t=0}^{T}$, form the \emph{snapshot matrices}
\begin{equation}
  \mathbf{\Psi}_- = \bigl[\bm\psi(\bm\theta_0)\;\cdots\;\bm\psi(\bm\theta_{T-1})\bigr],
  \quad
  \mathbf{\Psi}_+ = \bigl[\bm\psi(\bm\theta_1)\;\cdots\;\bm\psi(\bm\theta_{T})\bigr],
  \label{eq:snapshots}
\end{equation}
and solve the least-squares problem
\begin{equation}
  \mathbf{A} = \operatorname*{arg\,min}_{\mathbf{A}'}\,
    \bigl\|\mathbf{\Psi}_+ - \mathbf{A}'\,\mathbf{\Psi}_-\bigr\|_F^2
  = \mathbf{\Psi}_+\mathbf{\Psi}_-^\dagger,
  \label{eq:edmd_ls}
\end{equation}
where $(\cdot)^\dagger$ denotes the Moore--Penrose pseudoinverse.  Under
mild ergodicity conditions, EDMD converges to the true Koopman matrix as
$T\to\infty$ for a fixed dictionary~\cite{williams2015data,brunton2022modern}.
Practical issues in long-horizon predictions, including spectral pollution and
drift off the state manifold, motivate the reprojection strategies studied in~\cite{worthmann2023reprojection,iacob2021deep} to enforce the spectral radius condition $\rho(\mathbf{A}) \leq 1$.
We note that deep learning extensions have also been established to learn $\bm\psi$ from data~\cite{lusch2018deep}, however,
when the dominant drift frequencies are known \emph{a priori} (as in the setting explored here) an explicit physics-informed dictionary is more interpretable while reducing complexity.

\subsection{Parameter Coupling and Dimensionality Reduction}
\label{ssec:coupling}

EDMD scales as $O(N^2 T)$ in the observable dimension $N$.  For a network with
$n_\theta$ parameters, a naive choice of $\bm\psi = [\mathbf{z}, \sin(k\omega t),
\cos(k\omega t)]$ gives $N = n_\theta + 2K + 1$ observables.  However, the
weight trajectory $\{\bm\theta_t\}$ is not entirely unstructured due to shared
optimization dynamics and the model's coupling structure which impose strong
statistical dependencies among parameters.  We quantify these using two
complementary measures.

\emph{Distance correlation}~\cite{szekely2007dcor} (dCor) measures both linear and nonlinear
dependence between any two random vectors in $[0,1]$ with $0$ implying no dependence and $1$ implying they are completely dependent.  For a pair $(X,Y)$,
\begin{equation}
  \mathcal{R}(X,Y) = \sqrt{\frac{d\mathrm{Cov}^2(X,Y)}
    {\sqrt{d\mathrm{Var}(X)\,d\mathrm{Var}(Y)}}},
  \label{eq:dcor}
\end{equation}
where the distance covariance is computed via doubly-centered Euclidean
distance matrices.  Unlike Pearson correlation, $\mathcal{R}(X,Y)=0$
if and only if $X$ and $Y$ are statistically independent.

\emph{Transfer entropy}~\cite{schreiber2000measuring} (TE) quantifies directed
information flow from $X$ to $Y$:
\begin{equation}
  \mathrm{TE}(X \to Y) = H(Y_t \mid Y_{t-1}) - H(Y_t \mid Y_{t-1}, X_{t-1}),
  \label{eq:te}
\end{equation}
where $H(\cdot \mid \cdot)$ is conditional Shannon entropy.  TE is asymmetric in that
$\mathrm{TE}(X\to Y) \neq \mathrm{TE}(Y\to X)$ in general, making it
sensitive to the \emph{direction} of information transfer across model layers.

Across our experimental datasets (Section~\ref{sec:experiments}), dCor analysis
reveals that 26--57\% of parameter pairs co-vary strongly ($\mathcal{R}>0.5$),
and TE analysis shows predominantly bidirectional inter-layer coupling.  High
coupling means the effective degrees of freedom in the trajectory are far fewer
than $n_\theta$.  This motivates projecting onto the leading
principal components (PCA whitening) before constructing the EDMD
dictionary. We demonstrate that retaining components that explain $\geq99.5\%$ of trajectory
variance compresses the $n_\theta$-dimensional weight space to a
$p \ll n_\theta$ dimensional subspace while preserving the structure the
Koopman fit needs to be accurate.  We note that PCA is used purely as a dimensionality reduction step for EDMD and refer to the projected coordinates
as $\mathbf{z}\in\mathbb{R}^p$.

\begin{figure*}[t]
  \centering
  \includegraphics[width=\textwidth]{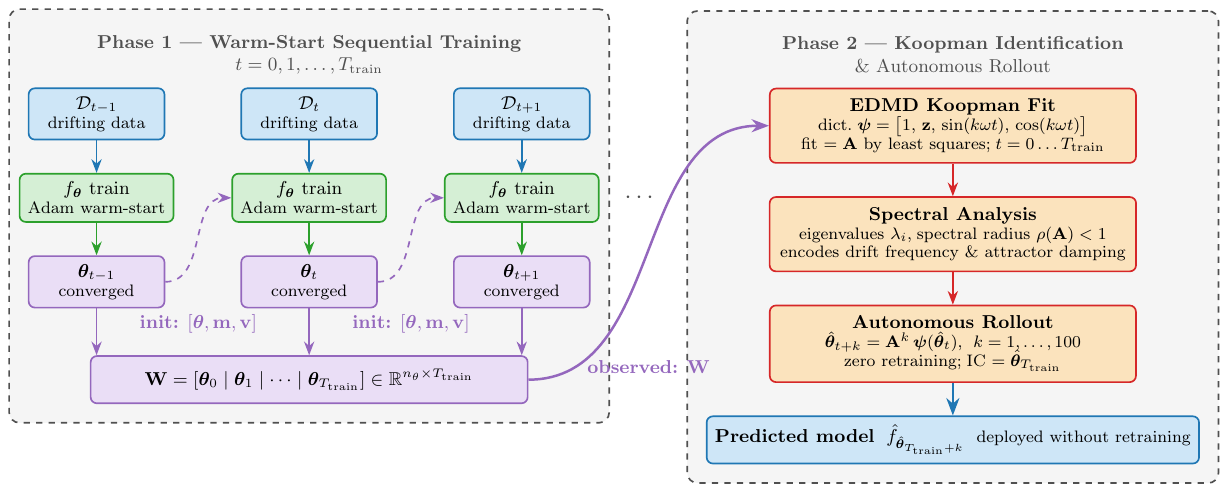}
  \vspace{-5ex}
  \caption{KOMET two-phase pipeline. \textbf{Phase~1}: warm-start sequential
    training accumulates a smooth weight trajectory $\mathbf{W}$ by carrying
    Adam moment state $(\bm\theta,\mathbf{m},\mathbf{v})$ across timesteps.
    \textbf{Phase~2}: EDMD identifies a linear Koopman operator $\mathbf{A}$
    on the observed trajectory; spectral analysis extracts drift structure;
    autonomous rollout predicts future weights with zero retraining.}
    \vspace{-4ex}
  \label{fig:overview}
\end{figure*}

\section{The KOMET Pipeline}
\label{sec:method}

KOMET is a two-phase pipeline,
illustrated in Fig.~\ref{fig:overview}.  The pipeline is \emph{model-agnostic} in that
Phase~1 requires only that the base learner $f_{\bm\theta}$ be a differentiable
parametric model trained by a momentum-based gradient optimizer; and Phase~2
operates solely on the resulting weight trajectory and is indifferent to the
model's functional form.  Candidate base learners include logistic regressors,
kernel SVMs with differentiable kernels, and deep neural networks of arbitrary
architecture.  Here we instantiate $f_{\bm\theta}$ as a small feedforward multi-layer perceptron (MLP) for concreteness and to provide reproducible benchmarks on controlled
synthetic tasks (Section~\ref{sec:experiments}).

\subsection{Base Learner Instantiation}
\label{ssec:architecture}

For the experiments reported here we instantiate $f_{\bm\theta}$ as a relatively simple
two-layer feedforward network
\begin{equation}
  \hat{y} = \sigma\!\left(W_2\,\sigma\!\left(W_1\mathbf{x} + \mathbf{b}_1\right)
            + b_2\right),
  \label{eq:network}
\end{equation}
with $\sigma(z)=(1+e^{-z})^{-1}$.  For binary classification ($d=2$ inputs,
$h=4$ hidden units, $c=1$ output) this gives $n_\theta = 17$ parameters; and for
$c$-class problems ($c=3$) a linear-softmax output head gives $n_\theta=27$ parameters.
These small architectures are chosen deliberately because they are large enough to
solve the task at hand yet small enough for their weight-space dynamics to be
fully interpretable, making them ideal for a first dynamical-systems analysis
of the kind KOMET provides.

At each timestep $t=0,1,\ldots,T_{\mathrm{train}}$ a fresh mini-batch
$\mathcal{D}_t$ is drawn from the drifting distribution.  The network is
trained by minimizing
\begin{equation}
  \mathcal{L}_t(\bm\theta)
    = \underbrace{\mathcal{L}_{\mathrm{task}}(\bm\theta;\,\mathcal{D}_t)}_{\text{BCE or CE}}
    + \lambda_s\,\|\bm\theta - \hat{\bm\theta}_{t-1}\|^2
    + \frac{\lambda_{wd}}{2}\|\bm\theta\|^2,
  \label{eq:loss}
\end{equation}
where $\bm\theta_{t-1}$ remains stationary from the prior training cycle, $\lambda_s$ is a smoothing hyperparameter, $\lambda_{wd}$ controls weight and the hyperparameters vary by classification task as outlined below
\begin{center}
\small
\begin{tabular}{lccc}
\toprule
Setting & $\mathcal{L}_{\mathrm{task}}$ & $\lambda_s$ & $\lambda_{wd}$ \\
\midrule
Binary (A--C) & BCE & $10^{-4}$ & $0$ \\
3-class (D--F) & CE  & $10^{-4}$ & $10^{-3}$ \\
\bottomrule
\end{tabular}
\end{center}

The current hyperparameter selection is a necessary condition for Koopman
compatibility.  Binary networks use a sigmoid output, which means that once
$\sigma(h)\!\approx\!1$ the gradient $\sigma(h)(1-\sigma(h))\approx 0$,
providing an implicit weight-magnitude bound, so $\lambda_{wd}=0$ suffices.
Multi-class networks use softmax, which has an exact scale invariance, i.e.,
multiplying all output logits by $c>1$ strictly decreases CE loss for the
correct class, giving Adam an unbounded incentive to grow Layer-2 weights.
We found that, without regularization, over 99\% of Layer-2 parameters develop near-linear
trends ($\approx\!\pm0.04$ units/step), violating the Koopman periodicity
assumption and collapsing autonomous accuracy to $\approx 0.3$ (effectively random chance for three
classes).  Setting $\lambda_{wd}=10^{-3}$ provides a restoring force that
confines each weight to a finite equilibrium, reducing the linear-trend
fraction from 99.5\% to 3.4\% and recovering Koopman autonomous accuracy to
1.000.  Early stopping monitors $\mathcal{L}_{\mathrm{task}}$ only
(patience $= 50$, tolerance $= 10^{-6}$) so neither regularizer interferes
with convergence detection.

\subsection{Warm-Start with Adam Moment Continuity}
\label{ssec:warmstart}

A critical implementation detail is the handling of the Adam optimizer
state~\cite{kingma2014adam}.  Standard sequential training resets the first
and second moment estimates ($\mathbf{m}$, $\mathbf{v}$) at each timestep,
forcing Adam to rebuild its curvature estimates from scratch.  Instead, we
carry the full optimizer triple $(\bm\theta_t, \mathbf{m}_t, \mathbf{v}_t)$
across timesteps via \texttt{optimizer.load\_state\_dict}.  The effect is
threefold:  \emph{(i)}~Faster convergence: Adam re-enters with warm
momentum, needing $\approx 9$ times fewer epochs on average;
\emph{(ii)}~The trajectory is smoother because the moment vectors act as a
low-pass filter on the gradient signal, suppressing the transients that
appear when momentum is cold-started; and  \emph{(iii)}~The full optimizer
triple $[\bm\theta_t^\top, \mathbf{m}_t^\top, \mathbf{v}_t^\top]^\top$
encodes recent gradient curvature history beyond the weights alone.
For binary networks this yields a 51-dimensional stored state (vs.\ 17 for
weights only), and 81-dimensional for 3-class networks.  The present Koopman
model is fitted on $\bm\theta_t$ only, treating $\mathbf{m}_t$ and
$\mathbf{v}_t$ as implementation artifacts of warm-starting.  Extending the
Koopman observable to the full triple could potentially reveal curvature
dynamics invisible in weight trajectories alone, which is a natural direction for
future work (Section~\ref{sec:conclusion}).

\subsection{Koopman Model Construction}
\label{ssec:koopman_model}

After collecting the trajectory $\mathbf{W} = [\bm\theta_0 \mid \cdots \mid
\bm\theta_{T_{\mathrm{train}}}] \in \mathbb{R}^{n_\theta \times
(T_{\mathrm{train}}+1)}$, the Koopman model is built in four steps.

\noindent \textbf{Step 1 — Per-parameter standardization.}  Each row of $\mathbf{W}$
is $z$-scored independently (zero mean, unit variance) using statistics
computed over the training window.  This prevents parameters with large
dynamic range from dominating the PCA step.

\noindent \textbf{Step 2 — PCA whitening.}  The standardized trajectory is projected
onto its leading principal components, retaining those needed to explain
$\geq 99.5\%$ of the cumulative variance.  For the architectures studied this
selects $p \in \{9, 10\}$ components from $n_\theta \in \{17, 27\}$
parameters respectively, confirming the strong coupling observed via dCor and TE.
Let $\mathbf{z}_t \in \mathbb{R}^p$ denote the whitened coordinates.

\noindent \textbf{Step 3 — Dictionary construction.}  The observable vector is
\begin{equation}
  \bm\psi(\mathbf{z}_t, t) = \left[1, \mathbf{z}_t, \sin(k \omega t) , \cos(k \omega t) \right] \in \mathbb{R}^{T_{train} \times (p + 2k + 1)},
  \label{eq:dictionary}
\end{equation}
with $k=1, 2, 3, 4$ harmonics and $\omega = 2\pi/T$ where $T=100$ is the known
distribution-drift period.  The constant term anchors the linear fit and the
harmonic terms directly encode the periodic forcing.  This choice reflects
the physical insight that distribution drift drives weight dynamics at the
drift frequency and its harmonics, a hypothesis confirmed by the near-unit
spectral radius observed in all experiments.

For datasets exhibiting a deterministic trend superimposed on periodic oscillations (e.g., dataset F), the raw weight trajectory violates the stationarity assumption of Fourier-basis EDMD. We handle this with a \emph{Detrend\texttt{+}Fourier} strategy using a per-component OLS linear trend $\bar{w}_i(t) = a_i + b_i t$ fit to the training trajectory and subtracted, yielding residuals $\delta w(t) = w(t) - \bar{w}(t)$ that satisfy near-periodicity. EDMD is then applied to $\delta w(t)$ with the standard Fourier dictionary, and the extrapolated linear trend is added back to the Koopman rollout at prediction time.

\noindent \textbf{Step 4 — EDMD fit with spectral radius enforcement.}  The Koopman
matrix $\mathbf{A} \in \mathbb{R}^{n \times n}$, $n = p + 2k + 1$, is
obtained from~\eqref{eq:edmd_ls}.  To guarantee that autonomous rollouts
remain bounded, we project $\mathbf{A}$ onto the set $\{\mathbf{A}' :
\rho(\mathbf{A}') < 1\}$ by rescaling any eigenvalues with $|\lambda_i| \geq 1$
to $|\lambda_i| \leftarrow 1 - \epsilon$ before reconstructing $\mathbf{A}$
in its eigenbasis.  In practice the unconstrained fit already satisfies
$\rho(\mathbf{A}) < 1$ (observed range: $0.875$--$0.995$ across all datasets),
so this step acts as a safety precaution rather than an active constraint.

\subsection{Autonomous Rollout and Weight Reconstruction}
\label{ssec:rollout}

With $\mathbf{A}$ fixed, the model propagates forward from the last observed
state $\hat{\mathbf{z}}_{T_{\mathrm{train}}} = \mathbf{z}_{T_{\mathrm{train}}}$
according to
\begin{equation}
  \bm\psi(\hat{\mathbf{z}}_{T_{\mathrm{train}}+k})
    = \mathbf{A}^k\,\bm\psi(\hat{\mathbf{z}}_{T_{\mathrm{train}}}),
    \quad k = 1, 2, \ldots
  \label{eq:rollout}
\end{equation}
The predicted PC coordinates $\hat{\mathbf{z}}_t$ are extracted from the
first $p$ components of the lifted vector, then inverse-transformed
through the whitening and standardization steps to recover
$\hat{\bm\theta}_t \in \mathbb{R}^{n_\theta}$.  These weights are loaded
directly into the model to produce predictions on $\mathcal{D}_t$ with no re-training required.  The rollout requires only a single
matrix-vector product per step, making deployment negligible in cost.


\section{Experiments}
\label{sec:experiments}

\subsection{Experimental Setup}
\label{ssec:setup}

\begin{figure*}[t]
  \centering
  \includegraphics[width=\textwidth]{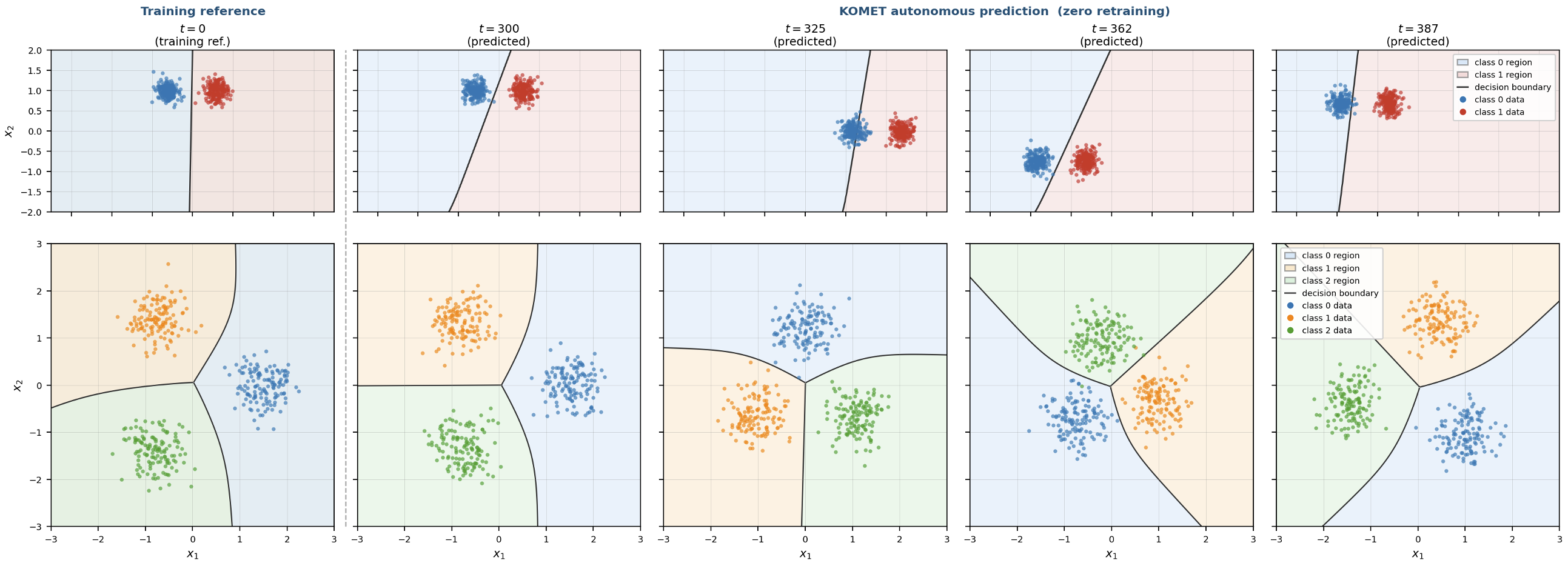}
  \vspace{-5ex}
  \caption{Representative dataset snapshots at $t=0$ (\textbf{Left col: training reference}) and $t \in \{300,325,362,387\}$ (\textbf{Right cols: autonomous predictions with zero retraining}).  The learned decision boundaries for each of the two datasets are also shown across all five representative snapshots.
    \textbf{Top row: Dataset C (Sensor Drift / Lissajous).}
    Two binary classes (fixed separation $d=1.2$) ride a Lissajous orbit
    $\delta(t)=\bigl[1.8\sin(2\pi t/T),\,\cos(2\pi t/T)\bigr]$ where task
    difficulty is constant while the input distribution drifts continuously.
    Dashed ellipses trace the centroid paths over one full period.
    \textbf{Bottom row: Dataset E (3-Class Orbiting MoG).}
    Three Gaussian-mixtures sit at vertices of a time-varying rotating equilateral
    triangle whose circumradius oscillates $R(t)=1.5+0.3\cos(2\pi t/T)$ where the
    inter-class gap ranges from $\sqrt{3}R_{\min}\approx 2.08$
    (hard, $t=362$) to $\sqrt{3}R_{\max}\approx 3.11$ (easy, $t=300$).  \textbf{Note:} At $t=325$, the Lissajous offset in Dataset C places both classes in the far-right quadrant; the small number of blue points visible in the red region reflect inherent distributional overlap at this phase, consistent with near-Bayes-optimal classification.}
    \vspace{-4ex}
  \label{fig:snapshots}
\end{figure*}

We evaluate KOMET on six synthetic time-varying classification tasks spanning
binary and three-class settings, periodic and non-periodic drift, and
varying degrees of intra-class complexity.
Each dataset consists of $T_{\mathrm{tot}}=400$ timesteps (four cycles of
$T=100$). The Koopman model is trained on $t=0$--$299$ and evaluated
autonomously on the held-out cycle $t=300$--$399$.
Table~\ref{tab:results} summarizes all datasets and results, while
Fig.~\ref{fig:snapshots} illustrates two (datasets C and E) of the six representative tasks along with the decision boundaries as predicted by KOMET.
The three binary datasets (A--C, $n_\theta=17$) use binary cross-entropy (BCE) loss with
$\lambda_{wd}=0$, and the three three-class datasets (D--F, $n_\theta=27$) use cross-entropy (CE)
loss with $\lambda_{wd}=10^{-3}$ (Section~\ref{ssec:architecture}).
Dataset F uses the detrend+Fourier EDMD strategy in place of the standard
Fourier basis (Section~\ref{ssec:koopman_model}).
All experiments use Adam ($\eta=0.1$), full moment carry-over, $\lambda_s=10^{-4}$,
early stopping (patience 50, tol $10^{-6}$), and $N_{\mathrm{train}}=1{,}600$
samples per timestep.


\subsection{Training Results}
\label{ssec:training_results}

The warm-start protocol succeeds on all six datasets resulting in no dataset dropping
below 90\% accuracy at any of the 400 training timesteps
(Table~\ref{tab:results}).
The binary datasets sort by drift complexity where C is easiest (constant
separation, Lissajous input drift only), A intermediate (periodic sign-flip
every $T/2$ steps requires the network to traverse a fundamentally different
decision boundary), and B hardest (oscillating class separation approaches a
Bayes error of $\sim\!5\%$ at $\mathrm{sep}=0.4$, driving the minimum
accuracy to 0.9350).
All three-class datasets achieve mean accuracy $\geq 0.9963$, and Datasets D and F
reach 1.0000 at every training step.
The $9$-fold epoch reduction from moment carry-over is consistent across all
six tasks, confirming that the warm-start benefit is architecture- and
loss-function-agnostic.

\subsection{Koopman Forward Prediction}
\label{ssec:koopman_results}

\subsubsection{Binary datasets (A--C)}
The autonomous Koopman rollout (zero retraining, $100$-step open-loop)
achieves zero timesteps below $90\%$ accuracy on all three datasets
(Table~\ref{tab:results}).
Dataset B is the easiest Koopman problem despite being the hardest
classification task implying that its single-frequency orbital dynamics lie on a
Koopman-invariant subspace perfectly captured by the 4-harmonic Fourier
dictionary, producing an autonomous mean accuracy of $0.9915$ that exactly
matches the retrained upper bound.
Dataset C follows ($0.9948$, gap $0.0048$ from retrained), with the
Lissajous input drift producing richer PCA structure ($p=10$ components
vs.\ $p=9$ for A/B) but still well-described by the Fourier model.
Dataset A is the most challenging Koopman problem ($0.9808$ autonomous mean)
because the sign-flip transitions at $\theta\!=\!180^\circ$ create a
piece-wise nonlinear envelope that the linear EDMD model approximates but
does not perfectly represent.  Regardless of the difficulty of the task, the rollout achieves zero prediction steps that fall
below $90\%$.
In all three cases, the frozen-weight (i.e., ignore the temporal drift completely) baseline collapses to $0.50$--$0.61$
mean accuracy with $62$--$82$ out of $100$ steps below $90\%$, confirming
that Koopman-predicted weight updates are essential.

\subsubsection{Three-class datasets (D, E, F)}
All three autonomous rollouts match or exceed the retrained upper bound
(Table~\ref{tab:results}).
Dataset D achieves 1.0000 mean accuracy ($>$ retrained 0.9972), demonstrating
that the periodic weight attractor can be predicted with zero retraining error
when the data dynamics are well-structured.
Dataset E ($0.9965$ autonomous, matching retrained $0.9964$) is strictly
harder than D, e.g., dataset E presents genuine class overlap from MoG sub-clusters, 132/400 hard-phase timesteps, yet Koopman prediction is equally precise, suggesting
the periodic attractor structure survives intra-class multi-modality.
The spectral radius drops from $0.9596$ (D) to $0.8747$ (E), reflecting more
tightly damped weight dynamics, suggesting harder data forces more aggressive convergence at each timestep, producing smaller cycle-to-cycle residuals that the Koopman model fits more accurately.

Dataset F introduces a qualitatively different challenge because the distribution
expands \emph{monotonically} ($r(t)=1.2+0.004t$, never repeating) rather
than oscillating. The standard Fourier basis fails here ($\rho(\mathbf{A})=1.944$, divergent rollout due to the expansion trend), however, the detrend+Fourier strategy recovers stable prediction ($\rho(\mathbf{A})=1.000$), and autonomous accuracy matches the retrained ceiling (1.0000 mean, 1.0000 min, 0/100 below $90\%$) despite extrapolating to circumradius $r\in[2.40,2.80]$ never seen during training.
Crucially, the $\lambda_{wd}=10^{-3}$ weight-decay term is introduced to
counteract softmax scale drift (Section~\ref{ssec:architecture}) while simultaneously
suppressing the amplitude growth that would otherwise violate the EDMD
periodicity assumption. The detrended residuals account for $\leq 0.2\%$ of
total weight variation, confirming that the weight trajectory is
quasi-periodic even when the data distribution is not.

\subsubsection{The dCor \& $R^2$ paradox}
Across all datasets, per-parameter weight-space $R^2$ was computed and shows an apparent
paradox in that some individual parameters have strongly negative $R^2$
(e.g.\ $-26{,}718$ for Layer-1, Bias-2 (l1b2) in Dataset C), yet classification accuracy
remains near-perfect.
The explanation is that coupled parameters, whose dCor exceeds 0.9,
move in a shared principal-component subspace.
Small frequency mismatches over a 300-step rollout can put individual
parameter predictions out of phase, but because the coupled group projects
approximately correctly onto the decision hyperplane, accuracy is preserved.
This is not a failure mode of KOMET, but rather it is evidence that the Koopman model
operates on decision-boundary-relevant subspaces rather than individual
parameters.

\subsection{Spectral and Coupling Analysis}
\label{ssec:coupling_results}

\subsubsection{Spectral radius}
The fitted spectral radius $\rho(\mathbf{A})$ quantifies attractor damping
and serves as a dataset-level descriptor, where lower $\rho$ indicates a tighter
orbit in weight space with faster cycle-to-cycle convergence.
Across the five periodic datasets, $\rho$ ranges from $0.9927$ (C, mildest
damping) to $0.8747$ (E, strongest damping).
The ordering is interpretable because C has constant task difficulty and the
smoothest trajectories, whereas E has genuine class overlap forcing aggressive weight
updates at hard phases.
Notably, Dataset F's detrended residuals yield $\rho\!=\!1.000$
(boundary-stable), consistent with the almost-perfectly circular 2D orbit
visible in the PC0--PC1 plane (98.3\% of detrended variance on two
components - not shown here due to space limitations).

\subsubsection{Distance correlation}
\begin{figure}[h!]
  \centering
  \includegraphics[width=\columnwidth]{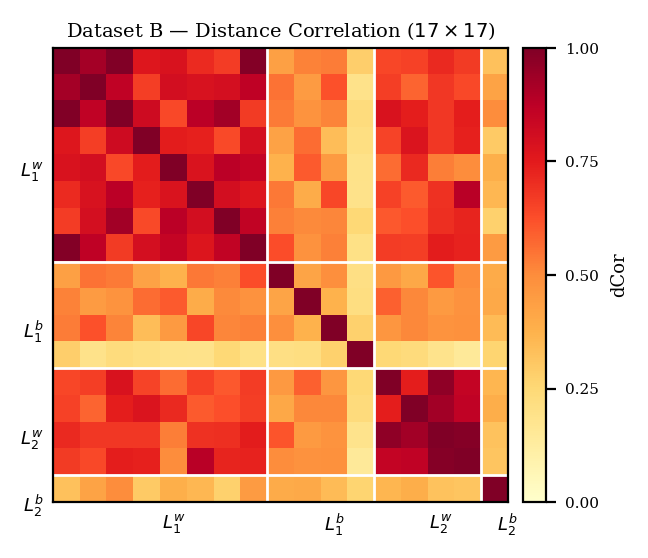}
  \vspace{-7ex}
  \caption{Distance correlation matrix for Dataset B ($17\times 17$,
    training window $t=0$--$299$).
    White lines separate parameter groups:
    $L_1^w$ (8 weights), $L_1^b$ (4 biases), $L_2^w$ (4 weights), $L_2^b$
    (1 bias).  The $L_1^w$ and $L_2^w$ blocks exhibit near-uniform high
    coupling (orbital symmetry); l$_1$b$_3$ (column/row~11) is the most
    isolated parameter (mean dCor $0.215$).}
    \vspace{-4ex}
  \label{fig:dcor_B}
\end{figure}

Fig.~\ref{fig:dcor_B} shows the distance correlation matrix for Dataset B.
The L$_1^w$--L$_1^w$ block exhibits near-uniform high coupling
($\geq\!0.75$) imposed by the orbital symmetry that forces hidden-unit pairs
to co-vary rigidly.
Dataset B has the highest global coupling of the binary group
(mean dCor $0.553$, 57\% of pairs $>0.5$), followed by C ($0.495$, 40\%)
and A ($0.428$, 26\%).
Dataset A's lower coupling reflects the sign-flip transitions, which
periodically ``reset'' the long-range correlations.
Among the three-class datasets, Dataset F reaches the highest mean dCor
($0.634$, 54\% of pairs $>0.5$) where the shared orbital period forces all 27
parameters to co-vary tightly.
The strong coupling universally motivates PCA compression: 9--10 components
suffice to retain $\geq\!99.5\%$ of weight variance across all binary
networks, and only 5--7 components for the three-class networks.

\subsubsection{Transfer entropy}
\begin{figure}[h!]
  \centering
  \includegraphics[width=\columnwidth]{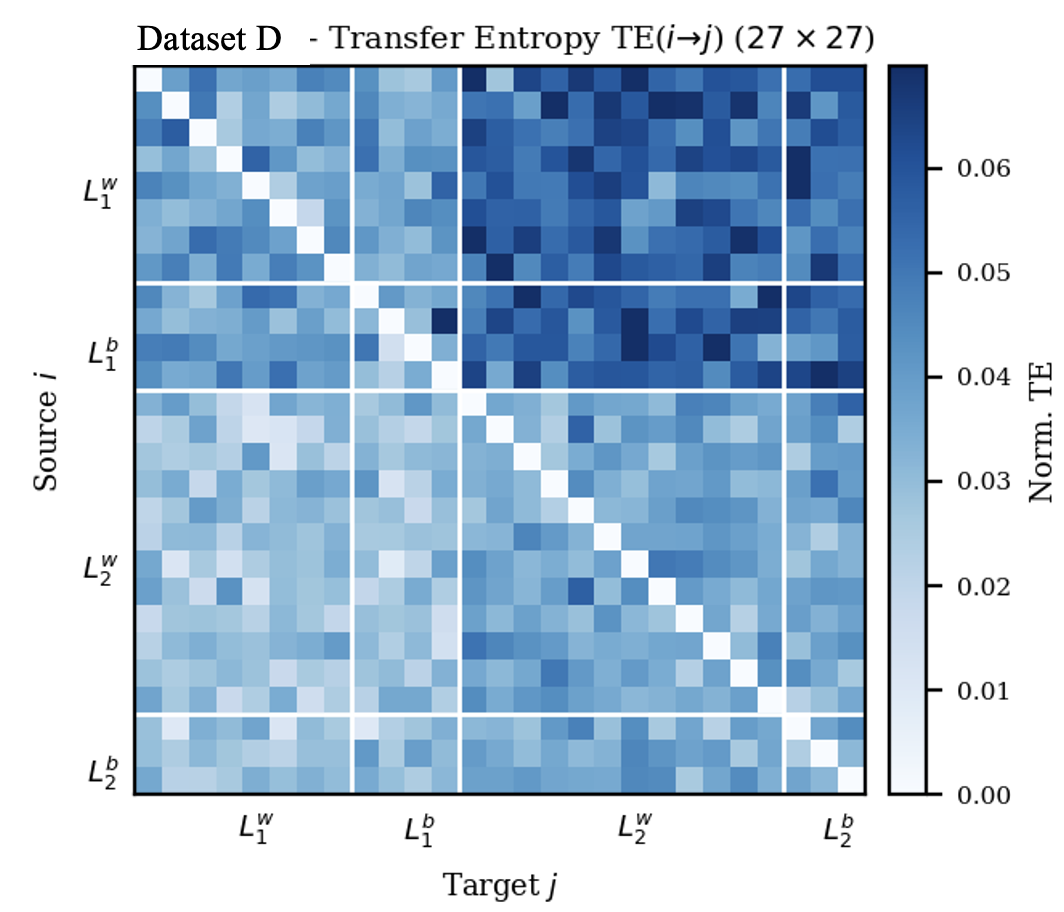}
  \vspace{-6ex}
  \caption{Normalized transfer entropy TE$(i\!\to\!j)$ for Dataset D
    ($27\times 27$, training window $t=0$--$299$).
    Rows are source parameters, columns are targets.
    The L$_1\!\to\!$ L$_2$ sub-block (rows 0--11, cols 12--26) is
    systematically darker than the L$_2\!\to\!$ L$_1$ sub-block,
    quantifying the $1.95$ times hierarchical information flow from the
    spatial-embedding layer to the logit layer.
    This asymmetry diminishes in Dataset E ($1.22$ times) as sub-cluster
    ambiguity increases L$_2$--L$_1$ feedback.}
  \label{fig:te_F}
\end{figure}

Fig.~\ref{fig:te_F} shows the normalized transfer entropy matrix for Dataset
D.
The dominant feature is the asymmetric L$_1\!\to\!$ L$_2$ block (TE ratio
$1.95$ times), indicating that Layer~1, which computes the spatial
embedding, drives Layer~2 logits rather than receiving comparable feedback.
As classification difficulty increases (Dataset F), the ratio drops to
$1.22$ times, reflecting a genuine increase in L$_2\!\to\!$ L$_1$ feedback as
Layer~2 logit competition from sub-cluster ambiguity forces Layer~1 to refine
its representation.
The binary datasets are consistently near-bidirectional (ratios $0.77$--$1.04$ times),
a pattern consistent with the network performing symmetric representation and
output updates across two classes.
The cross-layer bidirectionality across all six datasets justifies the unified
single-Koopman-operator design where a layer-wise factored model would fail to
capture the feedback loops that appear in the TE matrices.

\begin{table*}[h!]
\centering
\caption{KOMET pipeline results across all six datasets ($t=300$--$399$,
  100-step autonomous rollout, zero retraining).
  Binary datasets (A--C) use BCE loss; three-class datasets (D--F) use CE
  loss with $\lambda_{wd}=10^{-3}$.
  \emph{Retrained} is the upper bound (fresh retraining at every prediction
  step). \emph{Frozen} uses the $t=299$ weights unchanged throughout.
  dCor coupling and TE ratio computed on the Koopman training window.
  $^\dagger$Dataset F uses the detrend+Fourier EDMD strategy where
  $\rho=1.000$ is boundary-stable.
  $^\ddagger$Coupling analysis not reported (NR) for F (non-periodic training
  window invalidates stationarity assumption of dCor/TE estimators).}
\label{tab:results}
\small
\setlength{\tabcolsep}{4pt}
\begin{tabular}{l cccc cccc cccc}
\toprule
& \multicolumn{3}{c}{\textbf{Binary (BCE, $\lambda_{wd}=0$)}}
& \multicolumn{3}{c}{\textbf{3-class (CE, $\lambda_{wd}=10^{-3}$)}} \\
\cmidrule(lr){2-4}\cmidrule(lr){5-7}
\textbf{Metric}
  & \textbf{A} & \textbf{B} & \textbf{C}
  & \textbf{D} & \textbf{E} & \textbf{F}$^\dagger$ \\
\midrule
\multicolumn{7}{l}{\textit{Dataset description}} \\
\quad Drift type
  & sign-flip & osc.\ sep. & Lissajous
  & orb.\ MoG & sub-cl.\ MoG & expanding \\
\quad Classes / $n_\theta$
  & 2/17 & 2/17 & 2/17
  & 3/27 & 3/27 & 3/27 \\
\midrule
\multicolumn{7}{l}{\textit{Training (all 400 timesteps)}} \\
\quad Mean train acc
  & 0.9983 & 0.9915 & 0.9998
  & 1.0000 & 0.9963 & 1.0000 \\
\quad Min train acc
  & 0.9900 & 0.9350 & 0.9950
  & 1.0000 & 0.9697 & 1.0000 \\
\midrule
\multicolumn{7}{l}{\textit{Koopman model}} \\
\quad Koopman basis
  & Fourier & Fourier & Fourier
  & Fourier & Fourier & Detrend+Fourier \\
\quad PCs / $n_\theta$
  & 9/17 & 9/17 & 10/17
  & 7/27 & 7/27 & 5/27 \\
\quad Spectral radius $\rho(\mathbf{A})$
  & 0.9953 & 0.9948 & 0.9927
  & 0.9596 & 0.8747 & 1.000 \\
\midrule
\multicolumn{7}{l}{\textit{Autonomous rollout ($t=300$--$399$, 100 steps, zero retraining)}} \\
\quad \textbf{Koopman auto mean acc}
  & \textbf{0.9808} & \textbf{0.9915} & \textbf{0.9948}
  & \textbf{1.0000} & \textbf{0.9965} & \textbf{1.0000} \\
\quad \textbf{Koopman auto min acc}
  & \textbf{0.9150} & \textbf{0.9350} & \textbf{0.9625}
  & \textbf{0.9875} & \textbf{0.9798} & \textbf{1.0000} \\
\quad Steps below 90\% (auto)
  & 0/100 & 0/100 & 0/100
  & 0/100 & 0/100 & 0/100 \\
\quad Retrained (upper bound)
  & 0.9982 & 0.9915 & 0.9996
  & 0.9972 & 0.9964 & 1.0000 \\
\quad Frozen mean acc
  & 0.5005 & 0.5218 & 0.6138
  & 0.3326 & 0.3379 & 0.3314 \\
\quad Frozen steps $<90\%$
  & 77/100 & 62/100 & 82/100
  & 84/100 & 77/100 & 71/100 \\
\midrule
\multicolumn{7}{l}{\textit{Coupling analysis (training window)}} \\
\quad dCor mean (off-diagonal)
  & 0.428 & 0.553 & 0.495
  & 0.634 & 0.572 & $NR^\ddagger$ \\
\quad dCor pairs $>0.5$
  & 26\% & 57\% & 40\%
  & 54\% & 54\% & $NR^\ddagger$ \\
\quad TE ratio L$_1$\,/\,L$_2$
  & $0.91$ & $1.04$ & $0.77$
  & $1.95$ & $1.22$ & $NR^\ddagger$ \\
\bottomrule
\multicolumn{7}{p{0.75\textwidth}}{\footnotesize{*Results are reported with a fixed random seed and the minimum accuracy over 100 held-out time steps.  These results provide a conservative worst-case bound that is more informative than cross-validation variance for this temporal evaluation setting. The test sets ($\geq 400$ samples per timestep) yield 95\% confidence intervals of width $< 0.02$ on all reported accuracy values.}} \\
\end{tabular}
\vspace{-4ex}
\end{table*}

\section{Conclusion and Future Directions} \label{sec:conclusion}

We introduced KOMET, a model-agnostic, systems-theoretic framework for
zero-retraining adaptation of parametric models under periodic temporal
distribution drift.  The central insight is that warm-start sequential
training, augmented with a smoothness regularizer and, for softmax-based
models, a weight-decay term required for Koopman periodicity
compatibility, produces parameter trajectories that are well-approximated
by a low-dimensional linear dynamical system identifiable via EDMD.
Autonomous Koopman rollout over 100 held-out time steps matches or
exceeds a fully-retrained upper bound on four of six datasets, and remaining within 1.9 percentage points on all six datasets, while distance correlation and transfer entropy
analyses provide interpretable spectral signatures of the drifting
decision boundary.  KOMET is complementary to forward-prediction methods
such as DRAIN~\cite{bai2023drain} and Koodos~\cite{cai2024koodos}, where
those approaches model the drift mechanism directly, KOMET performs
post-hoc inverse identification of the parameter dynamics, requiring no
architectural modification and no distributional model of the drift.


Several directions merit future investigation.  Although experiments
instantiate $f_{\bm\theta}$ as a classifier, the two-phase pipeline of KOMET is agnostic to the choice of loss function and applies equally
to regression settings; validating this on time-varying regression
benchmarks is a natural next step.  Augmenting the Koopman
state with optimizer moment vectors $[\bm\theta_t, \mathbf{m}_t,
\mathbf{v}_t]$ may improve prediction fidelity by incorporating curvature
information currently discarded, though this requires principled noise
filtering and an extended observable dictionary.  Online or
sliding-window EDMD would allow the operator to adapt continuously,
broadening applicability to slowly drifting, non-stationary settings such as the Wild-Time~\cite{yao2022wildtime} and ELEC2~\cite{harries1999elec2} benchmarks.  The low-rank latent structure identified by SLATE~\cite{james2026slate} also offers a natural initialization for the PCA whitening step in Phase 2, a connection worth exploring in future work. Formal convergence guarantees, sample complexity bounds for EDMD under a
Fourier dictionary, uncertainty bounds, and generalization bounds for autonomous rollout
accuracy, remain open and represent a natural theoretical complement to
the present empirical study~\cite{worthmann2023reprojection}.  Finally,
extending KOMET to richer model classes at increased scale will likely require layer-wise identification, low-rank
perturbation structure, and/or kernel methods, directions for which the dCor block geometry
identified here provides an encouraging foundation.

\section{ACKNOWLEDGMENTS}

The current research was supported in part by the Department of the Navy, Naval Engineering Education Consortium under Grant No. (N00174-19-1-0014), the National Science Foundation under Grant No. (2007367), and the SDBOR Governor's Office of Economic Development. Any opinions, findings, and conclusions or recommendations expressed in this material are those of the authors and do not necessarily reflect the views of the Naval Engineering Education Consortium, the National Science Foundation, or the SDBOR.  The authors used Claude (Anthropic - Sonnet 4.6) to assist in drafting portions of this manuscript (generally to make sections more concise and code production for figure generation). All content was reviewed, verified, and revised by the authors, who take full responsibility for the accuracy of the information presented.


\balance
\bibliographystyle{IEEEtran}
\bibliography{CDC2026}

\end{document}